\pdfoutput=1

\documentclass[11pt]{article}

\usepackage{ACL2023}

\usepackage{times}
\usepackage{latexsym}

\usepackage[T1]{fontenc}

\usepackage[utf8]{inputenc}

\usepackage{microtype}

\usepackage{inconsolata}

\usepackage[flushleft]{threeparttable}
\usepackage{xurl}
\usepackage{nert}

\newcommand{\ELL}[0]{\textsc{ell}\xspace}
\newcommand{\ENG}[0]{\textsc{eng}\xspace}
\usepackage{arydshln}
\usepackage[noorphans,vskip=0.8ex]{quoting}
\usepackage{booktabs}

\title{ELQA: A Corpus of Metalinguistic Questions and Answers about English}

\author{
  Shabnam Behzad
  \\
  Georgetown University
  \\
  \eml{shabnam@cs.georgetown.edu}
  \And
  Keisuke Sakaguchi
  \\
  Tohoku University
  \\
  \eml{keisuke.sakaguchi@tohoku.ac.jp}
  \AND
  Nathan Schneider 
  \\
  Georgetown University
  \\
  \eml{nathan.schneider@georgetown.edu}
  \And
  Amir Zeldes 
  \\
  Georgetown University
  \\
  \eml{amir.zeldes@georgetown.edu}
}

\begin{document}
\maketitle
\begin{abstract}
We present ELQA, a corpus of questions and answers in and about the English language. Collected from two online forums, the >70k questions (from English learners and others) cover wide-ranging topics including grammar, meaning, fluency, and etymology. The answers include descriptions of general properties of English vocabulary and grammar as well as explanations about specific (correct and incorrect) usage examples. Unlike most NLP datasets, this corpus is \textit{metalinguistic}---it consists of language about language. As such, it can facilitate investigations of the metalinguistic capabilities of NLU models, as well as educational applications in the language learning domain. To study this, we define a free-form question answering task on our dataset and conduct evaluations on multiple LLMs (Large Language Models) to analyze their capacity to generate metalinguistic answers.
\end{abstract}

\section{Introduction}
\finalversion{\nss{a couple more analysis items I'd like to look at: (1) for each criterion and best human, low human, GPT-3 FS, histograms of average score of the two annotators. (2) same but absolute value of difference between annotator scores. (3) rank correlation between pairs of annotators, computed for each batch of 25. (4) which is harder, ENG or ELL? - (1) but broken down by subcorpus}\nss{make an appendix with details on the human evaluation: screenshot, instructions to annotators, time if you measured that}}

Language is so powerful that it can be reflected back on itself. Statements like “In informal usage, a \textit{steep learning curve} means something that is difficult (and takes much effort) to learn”\ or “In some cases, an adjective has both -ic and -ical forms, with no difference in meaning”\ expressly concern linguistic inventories,
structures, and behaviors. In other words, they are \emph{metalinguistic}---they use language to discuss language \citep[cf.][]{wilson-2013-toward}. They may concern a particular instance of language use, or properties of a
language or speaker in general; either way, they are metalinguistic in making linguistic phenomena the
subject matter of a linguistic utterance. For the rest of this paper, the term \emph{metalanguage} is used for natural language
text in which natural language is also the subject matter.

While NLP models have become powerful at \emph{predicting} text in many settings, it remains to be seen whether such capability extends to metalanguage---where linguistic strings are not being deployed to contribute to the discourse with their normal denotations, but rather, are treated as entities with linguistic properties (e.g., grammar, meaning).
One way this can be explored is in a question answering framework, which requires suitable datasets, ideally based on questions that are realistic and paired with high-quality answers.

In this paper, we present a corpus of metalinguistic questions and answers about English. The corpus is collected and carefully processed from two Stack Exchange forum sites: \emph{English Language \& Usage} (\ENG) and \emph{English Language Learners} (\ELL). It covers more than 70k questions on numerous topics about English such as grammar, meaning, fluency, and etymology along with answers. Our corpus, ELQA (English Language Questions and Answers), can serve as a tool to facilitate metalinguistic studies. Moreover, since questions in ELQA cover a variety of topics in English, it can be used in the educational and English language learning domains.

\begin{figure*}
    \centering
    \begin{subfigure}[b]{0.53\textwidth}
        \centering
        \includegraphics[width=1\textwidth ,trim = 0cm 5cm 9.25cm 4cm, clip]{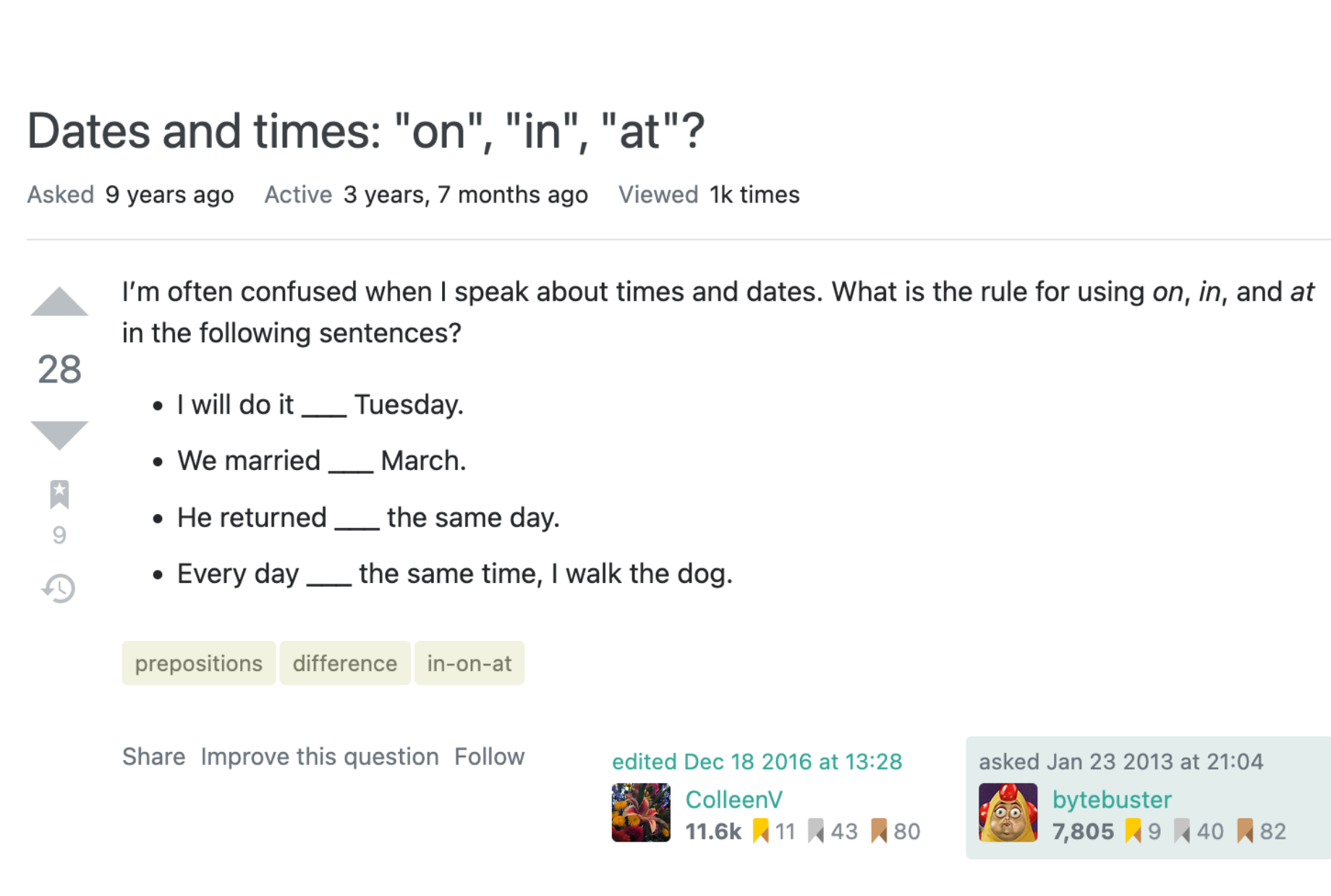}
        \caption{Question}
        \label{fig:q}
    \end{subfigure}
    \hfill
    \begin{subfigure}[b]{0.46\textwidth}
        \centering
        \includegraphics[width=1\textwidth , trim = 0cm 3.5cm 4cm 0cm, clip]{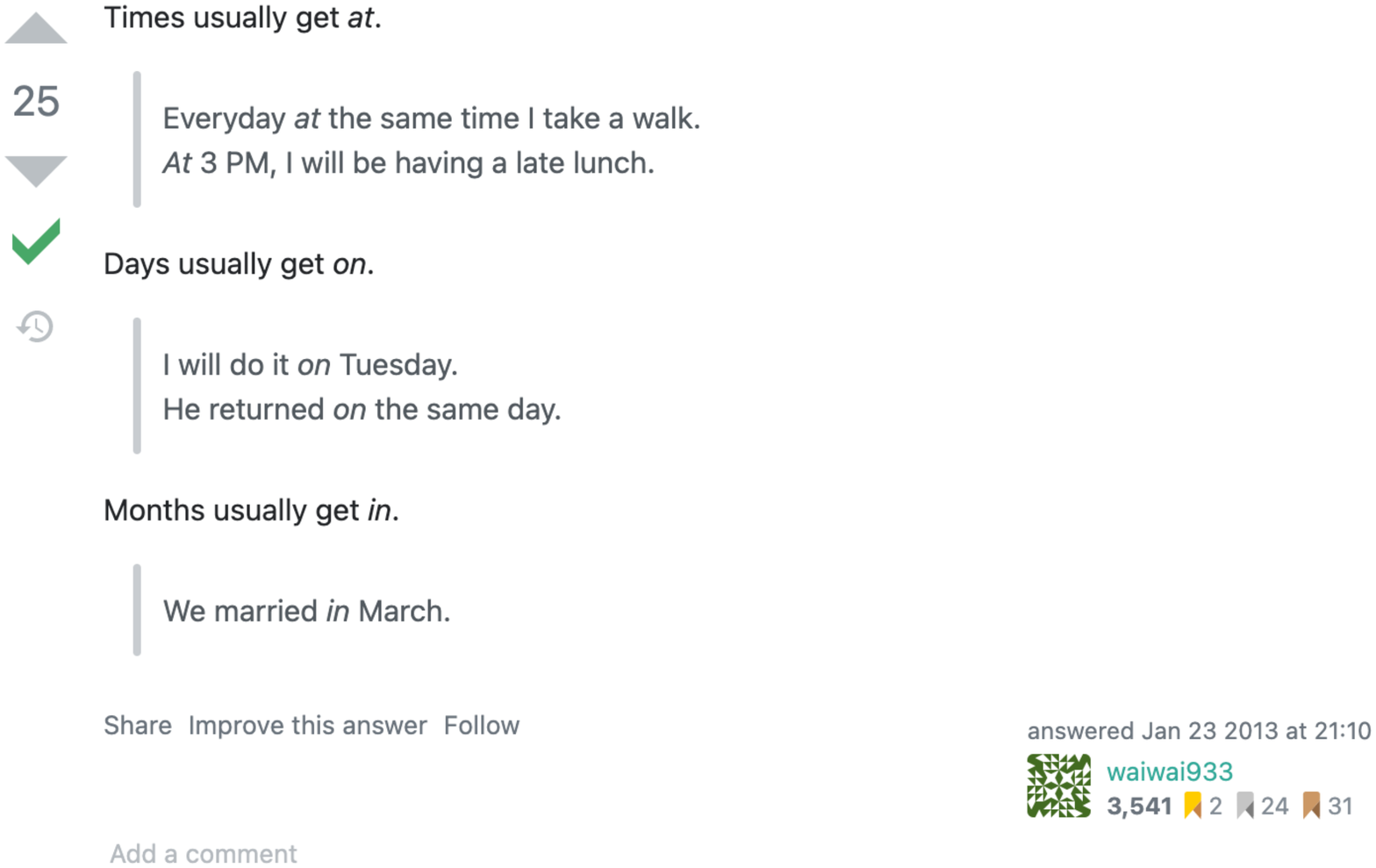}
        \caption{Answer}
        \label{fig:a}
    \end{subfigure}
        \vspace{0.1cm}
        \caption{Page screenshots from \ELL Stack Exchange.}
        \label{fig:ell-stackexchange}
\end{figure*}

As the first case study of ELQA,
we investigate the performance of current state-of-the-art NLP technology on free-form question answering in the English language domain. Additionally, we explore the possibility of building NLP models that can directly answer questions from language learners. We process a subset of ELQA and make it appropriate for this task. Then, we report on the results of both automatic and human evaluations using different experimental settings of T5\footnote{\url{https://github.com/google-research/t5x}} and GPT-3\footnote{\url{https://openai.com/blog/gpt-3-apps}} models. Although most of these models achieve high ratings for well-formedness, the validity of their answers is significantly lower than that of human-authored answers, indicating that this type of metalinguistic QA task is challenging even for large language models.

Our main contributions are: 1)~we release the first publicly available metalinguistic QA dataset,\footnote{\url{https://github.com/shabnam-b/ELQA}} focused on the English language; 2)~we present a taxonomy of questions in the corpus along with analysis; and 3)~we investigate to what extent LLMs are able to articulate appropriate generalizations about language in response to these questions.

\finalversion{The ELQA corpus and scripts are publicly available for future studies.\footnote{\url{https://github.com/shabnam-b/ELQA}}.}

\begin{figure}[t]
\centering
\includegraphics[scale=.6,trim = 7.5cm 2.5cm 2cm 3.5cm, clip]{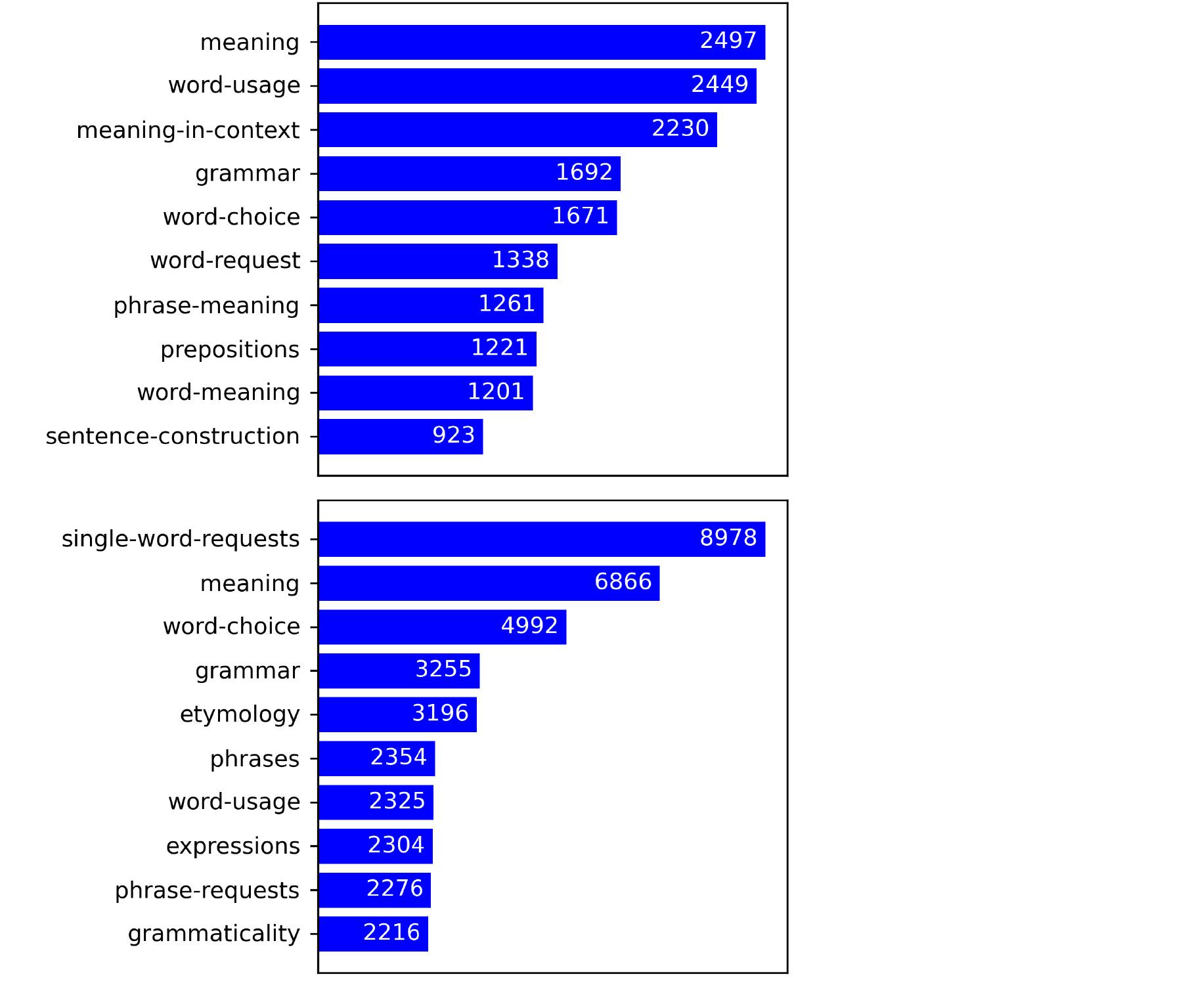}

\caption{Frequencies of top 10 user-assigned tags in \mbox{ELQA-large} subcorpora: \ELL (top) and \ENG (bottom).}
\label{fig:tag-stat}
\end{figure}

\section{Related Work}\label{sec:related}

Stack Exchange is a network of numerous CQA sites (originally and most famously, \emph{Stack Overflow}) built on a common platform. Stack Exchange forums have been featured in a number of previous datasets~\cite{yao2013want,hoogeveen2015,Ahmad2018ASO,penha2019mantis,campos-etal-2020-doqa,kumar-black-2020-clarq,rogers2021qa}, including the \emph{English} site (our \ENG) along with others such as \emph{Ask Ubuntu}, \emph{Android}, \emph{Gaming} and \emph{WordPress}~\cite{dos-santos-etal-2015-learning,nakov-etal-2017-semeval}. 
We focus on \ENG and \ELL as they concern the English language itself; we show that these datasets cover a wide range of metalinguistic questions.

Our use of these forums contrasts with previous work on metalanguage in corpora, which annotated and quantified mentions \citep{anderson2004types,wilson-2010-distinguishing,wilson2011search,wilson-2012-creation,Wilson2017}, but did not consider entire questions and answers about language.~\citet{TAYLOR2015127} studied metalanguage in online forums, but with a focus on the usage of metalinguistic expressions of mock politeness. More recently,~\citet{Bogetic2021} published the first corpus of contemporary Slovene, Croatian and Serbian media metalanguage texts.

So far, metalanguage has not been a focus in the QA domain---ours is the first publicly available English metalinguistic QA dataset. Most QA tasks are set up to have a question and a reference document, where the objective is to find the answer based on the document~\cite{fan-etal-2019-eli5,kwiatkowski-etal-2019-natural}. In this paper, we explored a type of “closed-book” question answering task~\cite{roberts-etal-2020-much,khashabi-etal-2021-gooaq-open}. To the best of our knowledge, this task has not been explored to date within the realm of English language questions that require significant generalization and adaptation rather than looking up facts.

\section{Constructing the Dataset}\label{lab:data}
\label{sec:data}

\begin{table}
\centering\small
\begin{tabular}{lcc}
\hline
\textbf{ELQA-large} & \textbf{\ELL} & \textbf{\ENG}\\
\hline
Total \# of Qs & \hphantom{0}23,520 &\hphantom{0}47,532 \\
Total \# of As & \hphantom{0}49,345 &152,315 \\

Avg.~Q length & \hphantom{0}92.41 & 102.41 \\
Avg.~A length & 158.25 & 137.90 \\\hdashline
Max.~A score & 392 & 581 \\
Min.~A score & $-$13 & $-$28 \\
Avg.~A score & 4.85 &5.15 \\\hdashline
Total \# of available tags & 513 & 951\\\hline
\textbf{ELQA-small} & \textbf{\ELL} & \textbf{\ENG}\\
\hline
Total \# of Qs & \hphantom{00}6,477 &\hphantom{0}14,234 \\
Total \# of As & \hphantom{0}18,389 &\hphantom{0}62,744 \\

Avg.~Q length & \hphantom{0}84.21 & \hphantom{0}89.25 \\
Avg.~A length & 156.29 & 118.66 \\\hdashline
Max.~A score & 392 & 581 \\
Min.~A score & $-$13 & $-$13 \\
Avg.~A score & 6.63 &6.73 \\\hdashline
Total \# of available tags & 437 & 823\\\hline

\end{tabular}

\caption{ELQA statistics on Qs (questions) and As (answers). To calculate average length in tokens, sequences were tokenized using SpaCy (\url{https://spacy.io/}).}
\label{tab:stat}
\end{table}

\begin{table*}[t]\centering 
\begin{threeparttable}
  \fontsize{8.2}{11.2}\selectfont
  \begin{tabular}{@{}>{\raggedright}p{0.15\linewidth} | p{0.2\linewidth} |p{0.59\linewidth}@{}}
  \hline
     \textbf{Question Type}& \textbf{Title} & \textbf{Body} \\\hline

    \textbf{Fluency} & “On my own way vs. “in my own way”? & Which one is correct <strong>in or on</strong> own way? <blockquote> <ul> <li>I usually help my closest friends on/in my own way.</li> </ul> </blockquote> \\
    \hline

    \textbf{Form to Meaning} & Wondering what "get by" means in this context & <blockquote> He tries to <strong>get by</strong> with the least amount of <strong>work possible</strong>. </blockquote> Could you tell me what this sentence means?\\
    \hline

    \textbf{Meaning to Form} & Grammatically correct synonym for "level of catastrophicness" &
    I'm trying to say something like this: <blockquote> We have developed a strategy to numerically rate the <strong>relative level of catastrophicness</strong> of a potential hardware failure. </blockquote> Looking at a thesaurus hasn't really helped me with this one. Can someone help me to convey this without using this ugly, incorrect grammar? \\
    \hline
    
    \textbf{Grammatical Analysis}& Should I modify a gerund using an adjective or an adverb?& I know that a gerund is a <strong>noun</strong>, so it should be modified by an <em>adjective</em>. However, it is also a <strong>verb form</strong>. Can I modify it by using an <em>adverb</em>?\\
    \hline

    \textbf{Other} &What is the etymology of 'physician'
 & I find myself confusing 'physician' and 'physicist' occasionally. While I know what they both mean, I am a little confused as to the use of 'physics' in 'physician'. How did the term 'physician' come to be used the way it is meant today? Lucky coincidence?\\
    \hline

  \end{tabular}
  \end{threeparttable}

  \caption{Example posts from \ELL and \ENG sites for different question types. (Original post URLs and author profile URLs are all available in the Appendix.)\finalversion{\nss{I notice that bolding and italics are shown elsewhere but not here. make it consistent and add a footnote explaining that some HTML tags are rendered}}}
  \label{tab:data-cat}
\end{table*}

We collect our data from two sites on Stack Exchange: \emph{English Language \& Usage} (\ENG)\footnote{\url{https://english.stackexchange.com/}} and \emph{English Language Learners} (\ELL)\footnote{\url{https://ell.stackexchange.com/}}. Sample screenshots of the site are shown in \cref{fig:ell-stackexchange}. The Stack Exchange data is publicly released under the CC-BY-SA~3.0 license. We preprocessed the data until 2021-12-06 collected from the Internet Archive\footnote{\url{https://archive.org/}} to be suitable for NLP studies and release it as ELQA. Additionally, some cleanup (e.g., removing posts marked as ``spam'' or ``offensive'') was done.
Fields for each entry (question) include the
title, body, user bio (if available), score (which is calculated based on up-votes and down-votes by other users), tags (user-assigned, related to the area/topic of the question), favorite count, and a list of answers. 
Textual content (body and user bio) is provided in two formats: HTML and plain text without HTML tags.

We release two versions of ELQA based on different preprocessing steps. In ELQA-large, we keep questions as long as they don't include any images (<img> HTML tag) and have an answer with a score of at least 2 (meaning at least two people other than the user posting the answer found it helpful). For ELQA-small, we applied further filtering to ensure that the data has the least amount of noise: a) questions should have a score of at least 2 (ensuring questions are clear and coherent), b) question has an answer with a score higher than 3 and c) there are no hyperlinks in at least one of the high-rated answers. The last step reduces noise and facilitates a fair comparison for the closed-book question-answering task~(\Cref{sec:gen}) with model-generated answers, as models cannot be expected to have access to the web to suggest valid URLs compared to humans who would search the web for appropriate resources to include in their answers.

For quality assurance, we also did a human annotation on ELQA-small. Two of the authors annotated 250 question and answer pairs for the following: 1) Is the question answerable? and 2) Does the answer fully address the question? We found 99.2\% of the questions answerable and 91.8\% of the answers acceptable. 

\Cref{tab:stat} contains overall statistics on both versions. \Cref{fig:tag-stat} shows the distribution of the 10 most common tags in each of the sites. Since users assign these tags to their questions (0 to multiple), similar or near-duplicate tags are common within the collection. Some form more general and more fine-grained variants, e.g.~`meaning' and `meaning-in-context'. In addition to available user-assigned tags, we manually inspected a large subset of the data to identify salient types of questions. These are defined below and illustrated in \cref{tab:data-cat}. We then labeled 100 random questions to get a rough estimate of their frequencies (two annotators annotated these 100 samples and they agreed on 92\% of cases in an overlapping subset).

\begin{itemize}[noitemsep, leftmargin=*, topsep=0pt]

    \item {\textbf{Fluency}~\textit{($\approx$38\% of questions)}:~Usually asking about a particular sentence, comparison of multiple sentences, and/or probing how an expression should be used in general. The user wants to know if X is correct, or to decide between multiple choices, which one is correct. “Correct” could mean grammatical, most natural\slash idiomatic, stylistically appropriate, conveying the intended meaning, etc. In Qs where options are provided by the user, there are cases in which 1) none of the choices are correct, 2) multiple choices are correct, and 3) only one is correct.}

    \item {\textbf{Form to Meaning (Interpretation)}~\textit{($\approx$19\% of questions)}:~Questions such as “What does X mean?”~(of an expression in general, or an encountered passage) or “What's the difference in meaning between X and Y?”.}

    \item {\textbf{Meaning to Form (Encoding})~\textit{($\approx$20\% of questions)}:~In these questions, the user gives some explanation/definition and asks for the term or for form to express it.}

    \item {\textbf{Grammatical Analysis}~\textit{($\approx$11\% of questions)}:~Questions about parts of speech and other aspects of syntactic analysis. (e.g.~“Is this a verb or an adjective?”; “Can an article ever go after the noun it modifies?”).
    Note that Fluency questions may mention grammatical terminology, but the grammatical categories are not the focus.}
    
    \item {\textbf{Other}~\textit{($\approx$10\% of questions)}:~Any other type of question not listed above. This includes questions about pronunciation, etymology, etc.}

\end{itemize}

As can be seen from the examples in \cref{tab:data-cat},
it is common for questions and answers to contain example usages, often visually distinguished with Markdown formatting (such as blockquotes, bullets, and italics) which we retain in the processed corpus markup. Examples can be incorporated into a post in a variety of ways---e.g., asking for an interpretation of one usage, as in the Form to Meaning example in \cref{tab:data-cat}, or contrasting multiple usages such as in the following question: 
\begin{quoting}\small
\noindent \textbf{Did VS Have done}\\ 
What is difference between the following statements: Did you tell your parents yet? Have you told your parents yet? Haven't you told your parents yet? Are these questions correct? why do we use one over another in some cases? What is the difference in meaning?
\end{quoting}

Usage examples provided in a question may be instances that the author encountered ``in the wild'' (such as in a novel or film), or in a grammar book or dictionary, or they may have been constructed by the user.
Answers sometimes include examples found through a corpus search.



\section{English Language Question Answering}\label{sec:gen}

\begin{table*}[t]\centering\small
\begin{threeparttable}
  \begin{tabular}{l|c|c|c|c|c}
     & ROUGE-1 & ROUGE-2 & ROUGE-L & BLEU &BERTScore\\\hline

GPT-3 FS &\textbf{31.2}&\textbf{8.5}& \textbf{20.3} & 10.8&\textbf{85.7} \\
GPT-3 FT-1000 &27.1 &7.0 &18.7 & \textbf{11.8}&85.2 \\

GPT-3 FT-100 & 25.6 & 6.0& 17.9& \hphantom{0}9.9&85.2 \\
T5-xxl & 28.1& 8.0&19.8 & \hphantom{0}4.7 &80.3 \\

T5-l & 21.2& 6.6& 17.7& \hphantom{0}4.1 &78.3\\
\hline
  \end{tabular}
  \end{threeparttable}

  \caption{Automatic evaluation scores (percentage) for different setups. The highest value in each column is bolded.}
  \label{tab:exp}
\end{table*}

\begin{table*}[t]\centering\small
\begin{threeparttable}
  \begin{tabular}{l|rr|rr|rr|rr|rr}
     & \multicolumn{2}{c|}{ROUGE-1} & \multicolumn{2}{c|}{ROUGE-2} & \multicolumn{2}{c|}{ROUGE-L} & \multicolumn{2}{c|}{BLEU} & \multicolumn{2}{c}{BERTScore}\\\hline

&\textbf{\ENG}&\textbf{\ELL}&\textbf{\ENG}&\textbf{\ELL}&\textbf{\ENG}&\textbf{\ELL}&\textbf{\ENG}&\textbf{\ELL}&\textbf{\ENG}&\textbf{\ELL}\\\hdashline
GPT-3 FS& 30.4 &32.8 & 8.0 &9.7 & 20.0&21.1&11.9&8.7&85.7&85.8 \\
GPT-3 FT-1000& 26.0& 29.6&6.3 &8.6 & 18.2&19.7&11.7&11.8& 85.2 &85.4\\
GPT-3 FT-100 & 24.8 & 28.0&5.4 & 7.3&17.6 &18.8&9.8&10.0&85.1 &85.2\\
T5-xxl & 26.8 &31.0 &7.1 &10.1 &19.1 &21.4&4.4&5.0&80.2& 80.4\\
T5-l &20.3& 23.2& 5.8& 8.3&17.1 &19.1&3.9&4.1&78.0 &79.0 \\
\hline
  \end{tabular}
  \end{threeparttable}

  \caption{Automatic evaluation scores (percentage) for different setups broken down by site}
  \label{tab:exp-sites}
\end{table*}

\begin{table*}
\centering\small
\begin{tabular}{l|cc|cc|cc|cc}
     & \multicolumn{4}{c|}{\textbf{C1}} & \multicolumn{4}{c}{\textbf{C2}}\\
 \textbf{Source}&Avg.~on \ENG&Avg.~on \ELL& Avg. &z& Avg.~on \ENG& Avg.~on \ELL&Total Avg. & z\\
\hline
Top-rated human & 4.81&4.87&4.83 & 0.34&4.44&4.57&4.49 & 0.64\\
Low-rated human &4.79&4.50&4.68 &0.15 &4.02&3.74&3.91 &0.28\\
GPT-3 FS &4.89&4.77&4.84 & 0.35&3.72&3.67&3.70 & 0.16\\
GPT-3 FT-1000 &4.50&4.43&4.47 &$-$0.07 &2.90&2.78&2.88 & $-$0.34\\
T5-xxl  &4.03&3.68&3.89 &$-$0.76 &2.17&2.78&2.25 & $-$0.74\\
\hline
\end{tabular}
\caption{Human evaluation scores (out of 5): average on each site (Avg.~on \ENG/\ELL), overall average (Avg.)\ and average z-score (z) on two criteria, C1 (well-formedness) and C2 (correctness\slash completeness), as described in the text.}
\vspace{-10pt}
\label{tab:human-study}
\end{table*}

\begin{table}
\centering\small
\begin{tabular}{l|cc|cc}
     & \multicolumn{2}{c|}{\textbf{C1}} & \multicolumn{2}{c}{\textbf{C2}}\\
 \textbf{Source}& \textbf{First} &\textbf{Last}& \textbf{First} & \textbf{Last}\\
\hline
Top-rated human &129 & \hphantom{0}9 & 104 & 10\\
Low-rated human &  114& 15& \hphantom{0}68& 20\\
GPT-3 FS & 131 & \hphantom{0}5 & \hphantom{0}68&30 \\
GPT-3 FT-1000 & \hphantom{0}97&28 & \hphantom{0}35 &62\\
T5-xxl  &  \hphantom{0}71&66 & \hphantom{0}23& 90\\
\hline
\end{tabular}
\caption{Number of times each system was ranked first (outright or tied) by an annotator, and the number of times it was ranked last (out of 150). }
\vspace{-10pt}
\label{tab:human-study-ranks}
\end{table}

\begin{table*}\centering
\begin{threeparttable}
  \fontsize{7.3}{9.3}\selectfont
  \begin{tabular}{@{}p{0.15\textwidth} | p{0.8\textwidth}}
  
    \hline

\textbf{Title}&\textbf{Question} \\\hdashline
    Explain the verb tense in "All that glitters is not gold"&I am confused by the maxim \textbf{All that glitters is not gold}. May I know why the word \textbf{is} is used there? In my known according the English grammar \textbf{glitters} is plural, So after plural we add verb+s. As in above example use auxiliary verb so why not \textbf{are} – why \textbf{is}?\\\hline

     \textbf{Source}& \textbf{Answer} \\\hline
    Top-rated human&Good question, but it's worth pointing out that \textit{glitters} in this context is a verb, not a noun. The subject of the sentence is \textit{all}. When \textit{all} is used as the subject of a sentence, it functions as a pronoun, and it can function as a singular or a plural pronoun. From NOAD <blockquote> \textbf{all} (\textit{pronoun}) [with clause ] the only thing (used for emphasis) : \textit{all I want is to be left alone}. (used to refer to surroundings or a situation in general) everything : \textit{all was well | it was all very strange}. </blockquote> So: <blockquote> \textit{\textbf{All} that glitters \textbf{is} not gold} </blockquote> is roughly equivalent to <blockquote> \textit{\textbf{Everything} that glitters \textbf{is} not gold} </blockquote> and the verb tense agrees. but: <blockquote> \textit{\textbf{All} who eat \textbf{are} not gluttons} </blockquote> is roughly equivalent to: <blockquote> \textit{\textbf{All the people} who eat \textbf{are} not gluttons} </blockquote> so there is no problem there, either.\\\hdashline
    
     Low-rated human&Grammatically is is used with third person singular and also s form of verb is used. "All" indicates a third person single group. \\\hdashline
     
     GPT-3 FS& The phrase \textit{all that glitters is not gold} is an idiom, which means that things that appear to be valuable may not actually be so. It is an old proverb, and the verb tense used in the phrase is an example of the so-called "historic present". The historic present is a literary device used to give a sense of immediacy to a narrative. It is used to describe events that happened in the past, but are being described as if they are happening in the present. In this case, the phrase is used to emphasize the timelessness of the proverb. So, while the verb \textit{glitters} is indeed plural, the verb \textit{is} is used in the historic present to give the phrase a sense of immediacy and timelessness.\\\hdashline
     
     GPT-3 FT-1000&The maxim is in the present tense, so the verb is also in the present tense: \textit{Glitters}. The past tense of \textit{glitter} is \textit{glittered}, and the past participle is \textit{glittered}. So if the maxim was in the past tense, it would be: <blockquote> All that glittered was not gold. </blockquote> The present tense maxim is a generalisation about something that is true in the speaker's experience. The past tense maxim would be a historical statement about what was true in the past.\\\hdashline
     
     T5-xxl& This is an interesting question, because it is not immediately obvious what the grammar rules are here. It is not a fixed phrase, and so subject-verb agreement must be based on the context. In this case, the \textit{that} is a prepositional phrase acting as a complement to the preposition \textit{all}. It could be rephrased as <blockquote> All that glitters is not gold </blockquote> It is true that  \textit{all} means "everything that glitters" - but it is equally true that  \textit{not everything that glitters} means "not everything that glitters is gold". It is much the same as <blockquote> Everything that glitters is not gold. </blockquote> The subject  \textit{all} is singular, so the verb must be singular as well -  \textit{is}. If it was \textit{twice} plural, then \textit{are} would be used, but it is not. It is interesting to note that Shakespeare used the same construction in \textit{Hamlet}: <blockquote> Everything that glitters is not gold. </blockquote> The subject \textit{that} is singular, and so the auxiliary verb \textit{is} agrees with the singular noun.\\

    \hline
    
    \end{tabular}
  
  \end{threeparttable}

  \caption{Example post with human and model-generated answers. More examples are available in the Appendix.}
  \label{tab:sample-gen-a}
\end{table*}


Large language models can produce output that is fluent and (at times) informationally adequate when presented with factual questions about entities in the world~\cite{roberts-etal-2020-much}. But how do such models perform when asked questions about the language itself? In this section, we investigate the free-form English language question answering task.

This task has the potential to benefit educational applications for language learners.
Research on NLP for educational purposes has investigated tasks such as automated grammatical error correction~\cite[\emph{inter alia}]{dale-etal-2012-hoo,ng-etal-2014-conll,bryant-etal-2019-bea,3474840WangSurvey}, question and quiz generation for language learning~\cite{sakaguchi-etal-2013-discriminative,chinkina-meurers-2017-question,marrese2018learning,vachev-etal-2021-generating-answer}, and automated essay scoring~\cite[\emph{inter alia}]{burstein2003rater,farag-etal-2018-neural}. Nevertheless, an application that has not been taken up by the educational NLP community is free-form question answering about language. Second language learners possess a degree of metalinguistic awareness about the language they are learning, and often turn to teachers or more advanced speakers with explicit questions about vocabulary, grammar, and usage. Community Question Answering (CQA) websites such as Stack Exchange have sites for language learners' questions and answers.
These sites require considerable effort by volunteers, and learners may have to wait for an answer---if an answer is provided at all. In fact, looking at the data from 2021-12-06 for \ENG and \ELL, 9\% of questions have no answers.

\subsection{Data}

We randomly divided ELQA-small into train/test/dev splits. This resulted in 21,175 Q\&A pairs in the train split and 3,107 Q\&A pairs in each of the dev and test splits. Answers in these splits have a score of at least 4. If there are multiple high-rated answers to a question, we include all of them for training. Some of these questions can be answered by looking at a dictionary or vocabulary list for descriptions. But many of them are explanations in relation to particular instances of language use and require significant reasoning rather than looking up facts. Thus in this setup, we do not have any external context\slash reference available at evaluation time, i.e.~this is a closed-book QA task.

The input for the task is \textit{Title: [Q title] <sep> Body: [Q body]}. We use the HTML version of ELQA for this task since metalinguistic mentions are usually distinguished via formatting (e.g., blockquotes, bullets) and the ultimate goal is a system that humans can easily use to get answers to their language-related questions.

\subsection{Setup}
We use T5~\cite{2020t5,roberts2022scaling} and GPT-3~\cite{brown2020language} as our models since they have been shown to be strong baselines in other QA domains. We believe the questions in ELQA offer new challenges for the QA task since they require different types of knowledge/understanding to be able to generate answers. Additionally, these questions contain noise (grammatical errors) and cases of textual metalanguage which is likely harder to comprehend for a model.

We fine-tune \textit{T5-l} and \textit{T5-xxl} for this task~\footnote{This took 5 days with v3-8 TPU (provided by Google)}. We saved multiple checkpoints during fine-tuning and evaluated them with the interpolation of BLEU~\cite{papineni-etal-2002-bleu}, BERTScore~\cite{bert-score} and ROUGE~\cite{lin-2004-rouge} on the dev set to choose the best-performing one (checkpoint at 75k updates, hyperparameters available in~\cref{tab:hyper} in the Appendix).

With GPT-3 we used \textit{text-davinci-003} and experimented with both fine-tuning (FT) on 100 and 1000 samples and a few-shot (FS) setting in which the model is given a few demonstrations of the questions and answers at inference time as conditioning, but no weights are updated~\cite{Radford2019LanguageMA}. In the FS setting, we show the model four Q\&A pairs since we wanted the model to see different question types but there were also limits on the input length. To select these 4 pairs, we randomly created 5 different sets of Q\&A pairs, evaluated on a subset of dev, and chose the best-performing set for the experiments (dev results available in Appendix, \Cref{tab:gpt3-dev}).

\subsection{Results}

\subsubsection{Automatic Evaluation}\label{sec:autoeval}

Results are shown in~\Cref{tab:exp}.~\textit{GPT-3 FS} outperforms all other methods in all metrics with a large margin except for BLEU Score. We also observed that using GPT-3 in a few-shot setup worked much better than the fine-tuned version. Looking at some of the model-generated answers, we noticed that the fine-tuned model tends to generate longer answers containing redundant text. We observed improvements when we used 1000 samples instead of 100 for fine-tuning and hence, fine-tuning on larger data might result in better performance, however, we only experimented with 100 and 1000 samples in this paper due to having limited resources.

Based on \cref{tab:exp}, \textit{T5-xxl} seems to perform similarly to \textit{GPT-3 FT-1000}. However, a small manual evaluation showed otherwise (\textit{GPT-3 FT-1000} answers were slightly better). Furthermore, we observe that the scores for even the best system are very low, but manual evaluations showed that the \textit{GPT-3 FS} generates fairly good answers in many cases. Due to these observations and also given the well-known limitations of automatic metrics for evaluating generation tasks~\citep{kasai2021bidimensional,Celikyilmaz2020EvaluationOT,bhakthavatsalam2021think}, we believe conducting human evaluation for deeper analysis is necessary for this task.

In~\cref{tab:exp-sites}, we show results for each site to see if one is more challenging than the other. Overall, models perform slightly better on \ELL based on automatic metrics---but we see in the next section (\cref{tab:human-study}) that there isn't really a meaningful difference between the sites when humans evaluate the answers.

\subsubsection{Human Evaluation}

Human evaluators were presented with the question title and body, and then asked to rate 5~answers: a top-rated human-provided answer, a low-rated human-provided answer, and answers generated by 3 of our best models:~\textit{GPT-3 FS, GPT3 FT-1000, T5-xxl}.

They were asked to give ratings (via a slider widget, on a 1--5 integer scale---the higher, the better) for two criteria (C1 \& C2):\footnote{The survey interface is illustrated in \cref{fig:survey} of Appendix~\ref{sec:more}.}
\begin{quoting}\small
\begin{enumerate}
    \item Does the answer look grammatically\slash structurally like a good answer (ignoring whether it answers the question)? 
    \item Is the information in this answer a valid response to the question (ignoring formatting\slash stylistic issues)?
\end{enumerate}
\end{quoting}
The first criterion aims to get a score for \textit{fluency and coherence} and the second one for \textit{correctness and completeness}.

We collected ratings for a set of 75 questions (375 different answers). Each question with its set of answers was evaluated by at least 2 raters, and then the average score was calculated based on their responses.\footnote{Evaluators consisted of 6 English native speakers who are senior NLP researchers and graduate students. The answer source was hidden and the order (5 answers) was randomized. Annotations took about 90 minutes on average.} We also report the average z-score which is calculated over each annotator's raw score distribution for each metric, intended to normalize interannotator variation in how the scale is interpreted for each of the two metrics  (details in \Cref{sec:zscore}).

The results of this study are shown in \cref{tab:human-study}. 
Overall, answers generated by \textit{GPT-3 FS} have a small gap with human answers in both C1 and C2\footnote{We selected half of the human evaluation samples from very recent posts (June 2021 until January 2023) on Stack Exchange and compared the results with older posts. The models' scores were comparable or better on the more recent data, so we didn't see evidence of models having an advantage due to the fact that they may have been trained on some of the data available on the web. For reference, human evaluation scores for recent data were Avg.~C1=4.82, Avg.~C2=3.83 and for older data, Avg.~C1=4.86, Avg. C2=3.61.}. \textit{GPT-3 FT-1000} comes next, with less accurate answers containing redundant text and hallucinations. The smallest model, \textit{T5-xxl}, ranks last.

Rankings based on human evaluations are available in~\Cref{tab:human-study-ranks}. These results are also indicating that model-generated answers are fluent in most cases, but they are not as good as human answers when correctness/completeness is considered (\textit{GPT-3 FS} is ranked first or as good as a top-rated human answer in only 45\% of cases).

\begin{figure*}[]
\minipage{0.32\textwidth}
  \includegraphics[scale=.31,trim = 3cm 8cm 3cm 8cm, clip]{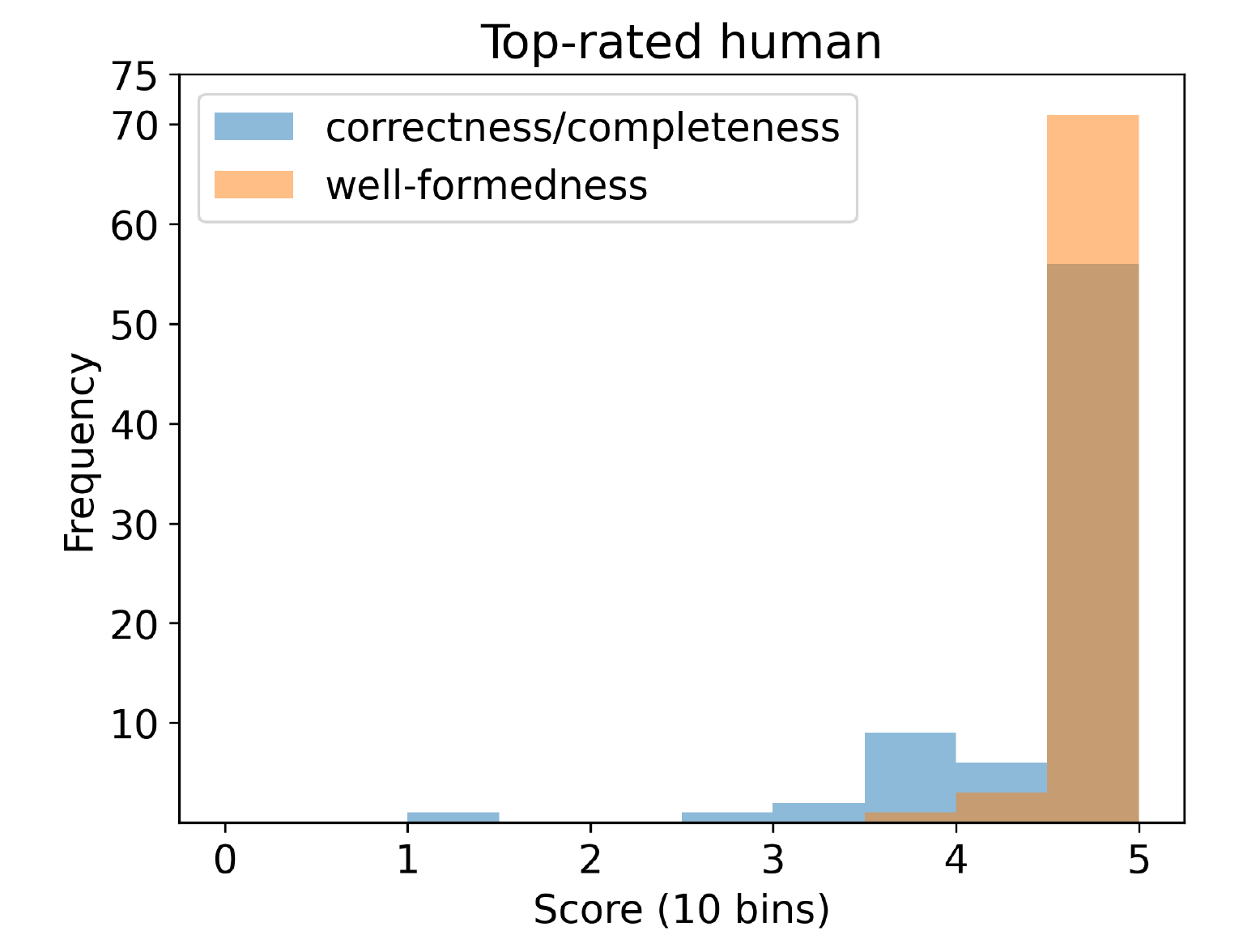}
  
\endminipage\hfill
\minipage{0.32\textwidth}
  \includegraphics[scale=.31,trim = 3cm 8cm 3cm 8cm, clip]{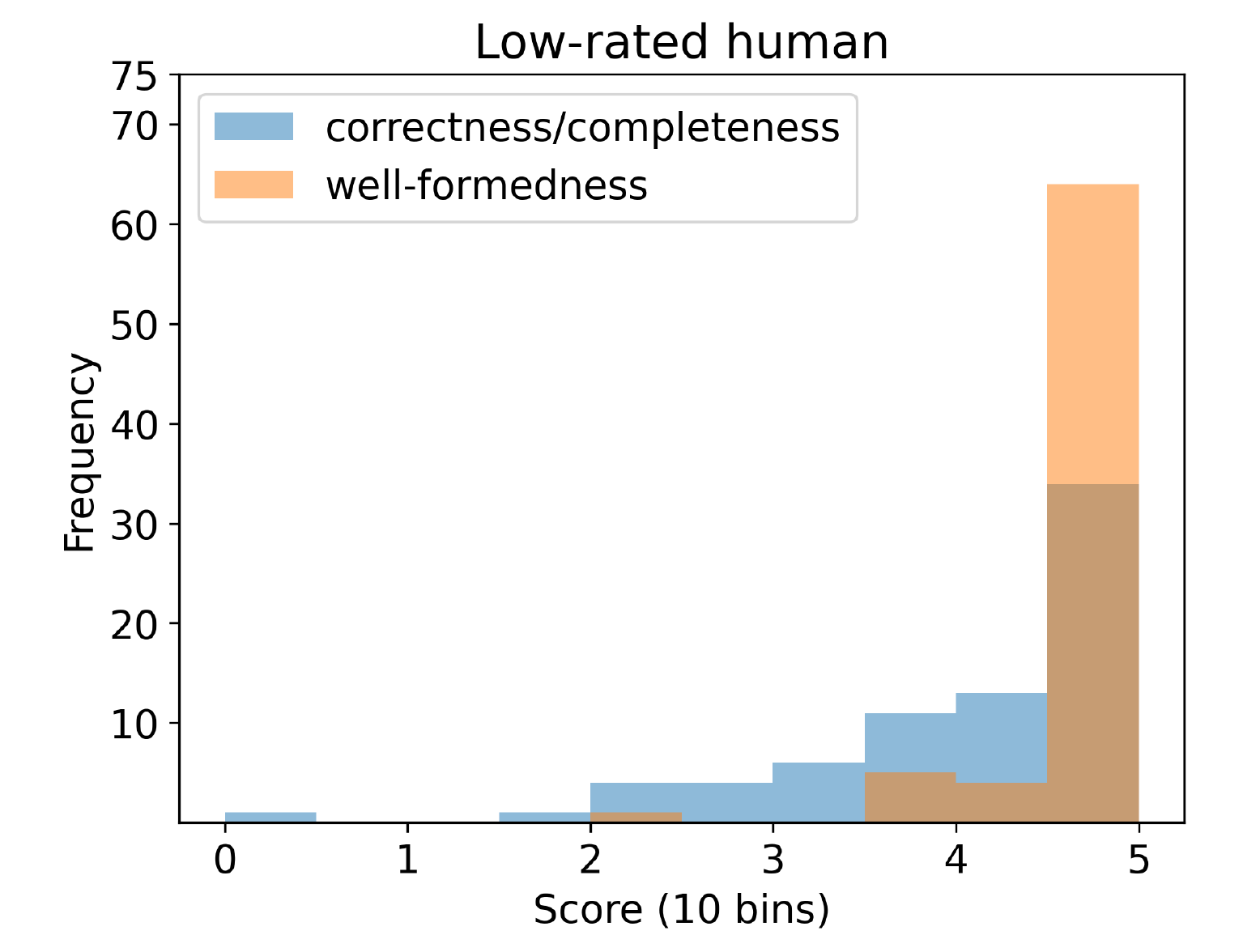}
\endminipage\hfill
\minipage{0.32\textwidth}%
  \includegraphics[scale=.31,trim = 3cm 8cm 3cm 8cm, clip]{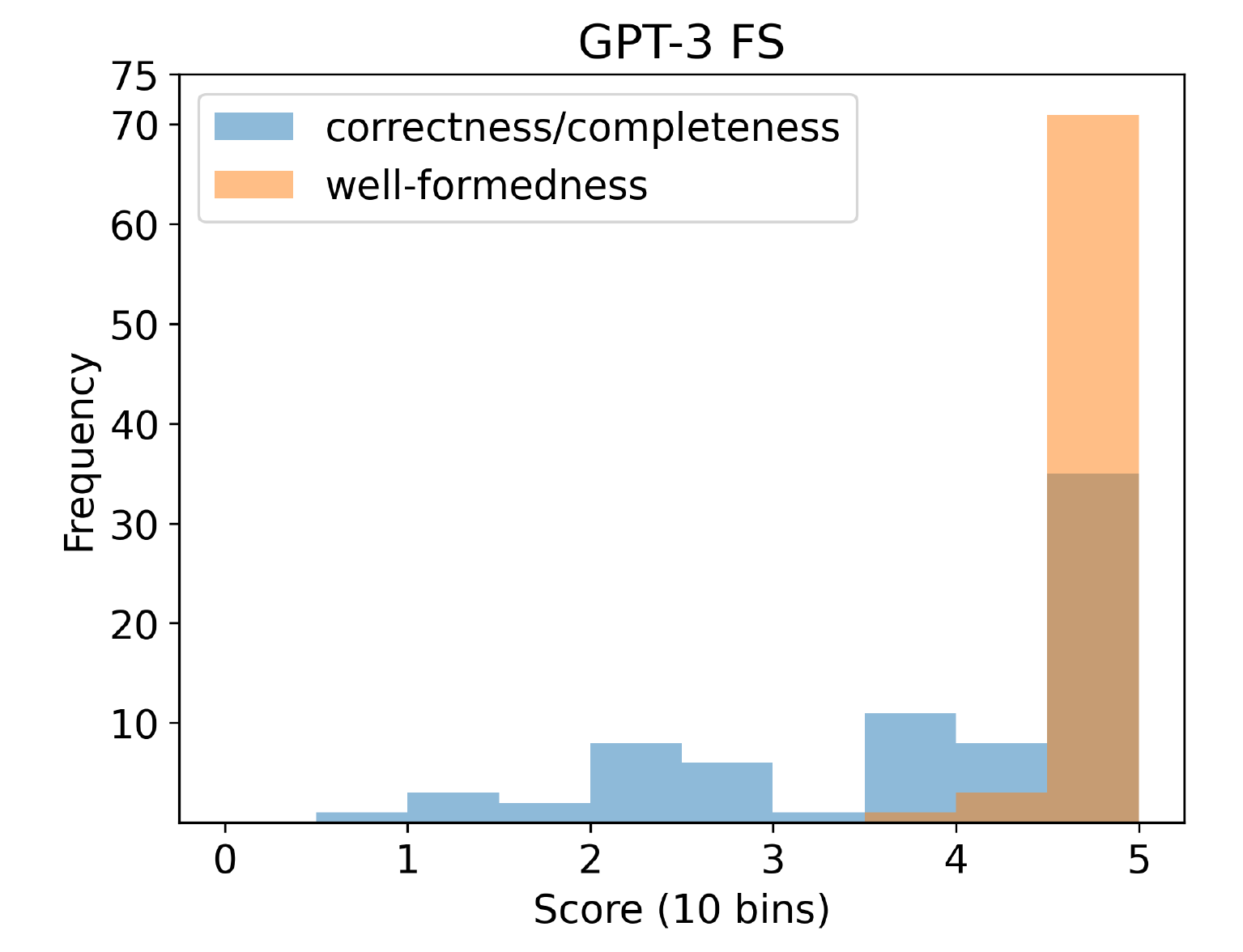}
  
\endminipage
\caption{Histograms of the ratings from our human evaluation of answers to 75~questions: the human-authored answer preferred by site users, the human-authored answer least preferred by site users, and our best model (GPT-3 FS). Each value is an average of two annotators' ratings. (Histograms for GPT-3 FT-1000 and T5-xxl are available in \cref{fig:more_human_charts} of  Appendix~\ref{sec:more}.)}\label{fig:human_charts}

\end{figure*}

For each criterion and Top-rated human, Low-rated human and \textit{GPT-3 FS}, histograms of the average score of the two annotators are plotted in \cref{fig:human_charts}. We can observe that GPT-3 FS and Low-rated human have very similar numbers of high-scoring answers (human evaluation scores), but the number of low-scoring human answers drops off gradually as quality decreases, while the distribution is more spread out for GPT-3 FS. I.e., the model has some moderately bad answers as well as some good ones, whereas Low-rated human answers cluster more on the upper end of the scale.

\paragraph{C1 (fluency/coherence).} All models generated fairly fluent and well-structured answers. We even notice that \textit{GPT-3 FS} scores higher in well-formedness than human answers. We looked at those samples and we believe there are two main reasons for this: 1)~Some human answers were very long, containing multiple different quotes from different sources. On average, our evaluators preferred the structure of answers from \textit{GPT-3 FS}, which took the form of a short paragraph addressing the question. 2)~Some human answers have a more casual/conversational nature. On the other hand, \textit{GPT-3 FS} generated more authoritative-sounding, teacher-like answers with complete sentences, which were preferred by our evaluators in some cases.

\paragraph{C2 (correctness/completeness).} On average, models are worse on this metric, though sometimes they did produce acceptable answers (perhaps because variants of the question are commonly discussed on the web).

One challenge in this domain is that questions, particularly from {\ELL}, may not be fully fluent. In the \textit{Fluency} example from~\Cref{tab:sample-gen-a} we see that there are some grammatical errors in the question that are unrelated to the topic of the question.
In addition, the questioner uses incorrect terminology, mentioning verb tense in the post title 
even though the question is actually about subject-verb agreement with respect to number (as can be inferred from the examples). While the good human response correctly flags the incorrect terminology and answers the underlying question, GPT-3 models give irrelevant answers about tense. 

Another correctness failure from \textit{GPT-3 FS} can be seen in the following pair involving a \textit{Meaning-to-Form} question:

\begin{quoting}\small
\noindent \textbf{What is the analog equivalent of ’digitally’?}\\ 
In electronics, we can solve a problem digitally or \textit{analoguely} ? That doesn't sound right, but surely there must be a word I can use? \textbf{Update}: some definitions: \textit{digitally:} Relating to or being a service that provides information expressed in discrete numerical form. \textit{analoguely:} Relating to or being a service that provides information expressed in continuous, non-discrete form. \textbf{(This is my made up definition of a word I made up)}. \textit{analogously:} Similar or alike in such a way as to permit the drawing of an analogy. As you can see, the word \textit{analogously} doesn't mean the opposite of \textit{digitally}, it means something else.\\
\textbf{Answer:} The analog equivalent of 'digitally' is 'analogously'. Analogously means "in a similar way" or "by comparison". It is used to describe a situation where two things are similar in some way, but not necessarily opposite. For example, you could say that a digital signal is analogous to an analog signal, because they are both signals, but they are different in terms of how they are represented.
\end{quoting}

We see that the model begins by ignoring the user's explanation that \emph{analogously} does not have the desired meaning, and suggests it anyway. 
The rest of the model's answer gives a (valid) general definition of \emph{analogously} and clumsily attempts to apply it to the two kinds of signals, which is not responsive to the question.
It may be the case that models particularly struggle with questions for which there is no straightforward solution (in this case, no adverb derived from \emph{analogue} has the user's intended meaning).

\subsection{Discussion}

\subsubsection{Metrics}

Our human evaluations found that the high-rated human answers are the best, and \textit{GPT-3 FS} is roughly on par with the low-rated human answer, while other models are worse.

As noted in \cref{sec:autoeval}, we did not find the automatic metrics very informative. We believe this is due to the long and free-form nature of this task: concepts can be explained in different ways and answers can include different examples which make automatic evaluations very challenging. For example, for the question \textit{Why is this sentence: "Additional nine features were added…" incorrect?} a human answer is:
\begin{quoting}\small
\noindent It has to do with the order of the adjectives. For example, consider this sentence: <blockquote> Happy nine men walk into a bar. </blockquote> Both \textit{nine} and \textit{happy} are adjectives, but we are really intending \textit{nine} to describe the \textit{happy men}, not \textit{happy} describing the \textit{nine men}. I don't .... 
\end{quoting}

Model-generated answers mostly mirror the examples in the question. But answers given by humans mention new examples---which are not in the question (and therefore unlikely to come up in other answers). Thus a good answer may mention superficially off-topic terms metalinguistically, which presents evaluation challenges unlike those of summarization or MT, for example~\cite{reiter-2018-structured,Celikyilmaz2020EvaluationOT}.
For QA tasks in metalinguistic domains such as ours, caution is warranted when applying automatic metrics that are unaware of the use/mention distinction. Human evaluation is likely to be more illuminating. We release our full human evaluation data to enable detailed comparisons in future research.

\subsubsection{Qualitative Analysis}

We examined behavioral patterns for the best model, \textit{GPT-3 FS}, organized by question type, with observations as follows.

\paragraph{Interpretation and Encoding:} These are the categories where the model performs the best, especially if the question has a straightforward answer (for example, there aren't several possible answers based on context). Topics in these categories often center on definitions of vocabulary and idioms. 
These questions may involve less reasoning to answer, and models may be leveraging content from the online dictionaries and thesauri in their web training data.

\paragraph{Grammatical Analysis:} When asked about general concepts and rules of grammar, the model-generated answers are usually acceptable:

\begin{quoting}\small
\noindent \textbf{-ed and -ing adjectives}\\ 
What is the difference? <blockquote> I am interesting in mathematics. </blockquote> versus <blockquote> I am interested in mathematics. </blockquote> Murphy intermediate grammar Unit 97 tells me that I cannot use the former. Extended and general explanations will be appreciated.\\
\textbf{Answer:} The difference between \textit{-ed} and \textit{-ing} adjectives is that the \textit{-ed} adjectives describe a feeling or emotion that is felt by the subject, while the \textit{-ing} adjectives describe a quality of the subject itself. In the example sentence, \textit{interested} is the correct word to use because it describes the feeling of interest that the subject has in mathematics. \textit{Interesting} would be incorrect because it describes the quality of the subject itself, not the feeling of the subject. For example, if you were to say "This book is interesting," you would be describing the book itself, not the feeling you have towards the book. On the other hand, if you were to say "I am interested in this book," you would be describing the feeling you have towards the book.
\end{quoting}

However, when questions become more specific to examples, sometimes the model struggles:

\begin{quoting}\small
\noindent \textbf{What does "in" refer to?}\\ 
<blockquote> A symbiotic relationship is an interaction between two or more species \textbf{in} which one species lives in or on another species.\\
\textbf{Answer:} It refers to the \textit{other} species. The symbiotic species lives \textbf{in} the other species.
\end{quoting}

The example in the question uses the word \emph{in} twice. Rather than addressing the first token, which the questioner has bolded, the answer addresses the second one (denoting the physical relationship between one species and another).

\paragraph{Fluency:} The model correctly answers many of the fluency questions. Others require reasoning and understanding of context, and the model struggles to generate a correct answer. For example:
\begin{quoting}\small
\noindent \textbf{Returning for the nth time}\\ 
Suppose I went to the supermarket three times. Is my third trip considered my "second time returning there" or my "third time returning there?" Thanks. (I know this question sounds silly, but English is my second language...)\\
\textbf{Answer:} Your third trip to the supermarket would be considered your \textit{third time returning there}. The phrase \textit{returning for the nth time} is used to refer to the \textit{nth} time you have returned to a place, not the \textit{n-1}th time.
\end{quoting}

We also observe that when the correct usage of a case is not mentioned in the question itself, the model sometimes suggests incorrect corrections and wrong explanations:

\begin{quoting}\small
\noindent \textbf{Why is this usage of "to" incorrect?}\\ 
I am in communication with a friend who is learning English. I noticed her using "to" in the wrong context, but I'm unable to explain which rule she is breaking. <blockquote> I can't access \textbf{to} Gmail. </blockquote>\\
\textbf{Answer:} In this sentence, the word \textit{to} is incorrect because it is being used as a preposition. The correct word to use in this context is \textit{for}. <blockquote> I can't access \textbf{for} Gmail. </blockquote> The word \textit{to} is used as a preposition when it is followed by a verb, such as \textit{go}, \textit{come}, \textit{run}, etc. For example: <blockquote> I am going \textbf{to} the store. </blockquote> However, when \textit{to} is followed by a noun, it is usually incorrect. In this case, the correct word to use is \textit{for}. <blockquote> I need a gift \textbf{for} my friend.</blockquote>
\end{quoting}

\subsubsection{Outlook}

 Overall, we were surprised by the quality of many of the answers from \textit{GPT-3 FS}: many would likely have received high user ratings if submitted as answers on the site. At the same time, the model is not to the point where we would want to trust its answers without human review. We believe that answer confidence estimation---so that users can be shown only the best model-generated answers---is an important direction for using learner QA models in practice~\cite{10.1162_tacl_a_00407}.

\section{Conclusion}

We presented ELQA, a dataset containing metalinguistic questions and answers about the English language. We provided analysis and a taxonomy of the data, along with experiments on \textit{free-form answer generation} and investigated the extent to which language models can articulate their generalizations about language. Since many of the questions in ELQA were asked by language learners, it forms a potentially useful and so far untapped resource for educational NLP purposes and metalinguistic question answering.
We release the dataset to enable further studies of this task.

\section*{Ethics Statement} We have released a processed version of an already public online forum dataset, in a manner consistent with the terms of the license, which require attribution of all posts (\cref{sec:data}).
\noindent
The models we have presented are intended only as baselines for future research, not for deployment. Models should be carefully stress-tested for undesirable heuristics\slash biases before deployment. 
Systems for the generation task, in particular, would risk misleading language learners with plausible but incorrect answers, so it is important to not deploy a generation system until it is approximately as reliable as existing non-automated alternatives, and to present the output with caveats. Potential biases reflecting the demographics of authors represented in the training data (in terms of native language, level of English proficiency, etc.)\ also need to be considered if models are deployed for different target populations.

\section*{Limitations}

One limitation of our dataset, ELQA, is that the corpus only contains questions in English and about English. However, Stack Exchange has sites with questions about other languages and our main data extraction scripts are general enough that they can be used to create corpora for other sites on Stack Exchange. Of course, language-specific processing steps, quality assurance and analysis must be applied before releasing such data.

Most importantly, the models we have presented here are intended only as baselines for future research, not for deployment. Potential biases reflecting the demographics of authors represented in the training data (in terms of native language, level of English proficiency, etc.)\ also need to be considered if models are deployed for different target populations. Moreover, many of these types of questions are found on the web, and a lot of the same topics are brought up by many users, so a model's ability to generate correct answers cannot necessarily be attributed to abstract reasoning.

\section*{Acknowledgements}
We thank the anonymous reviewers for their insightful comments. We thank Daniel Khashabi for helpful discussions and feedback. This research was supported in part by NSF award IIS-2144881.

\bibliography{acl2023}
\bibliographystyle{acl_natbib}

\clearpage
\appendix

\section{Data Credits}

The Stack Exchange license requires that any Internet use of the content should include a hyperlink directly to the original question and the profile of the authors. Below are URLs for all the examples used in this paper. The post URL incorporates the post title.

\begin{itemize}[noitemsep, leftmargin=*, topsep=10pt]
\small
    \item \url{https://ell.stackexchange.com/questions/12/dates-and-times-on-in-at} (Q by \href{https://ell.stackexchange.com/users/27/bytebuster}{bytebuster}, A by \href{https://ell.stackexchange.com/users/33/waiwai933}{waiwai933}) 
    
    \item \url{https://ell.stackexchange.com/questions/146633/on-my-own-way-vs-in-my-own-way} (Q by \href{https://ell.stackexchange.com/users/56970/bavyan-yaldo}{bavyan-yaldo})
    
    \item \url{https://ell.stackexchange.com/questions/19684/wondering-what-get-by-means-in-this-context} (Q by \href{https://ell.stackexchange.com/users/3751/nima}{nima})
    
    \item \url{https://english.stackexchange.com/questions/74896/grammatically-correct-synonym-for-level-of-catastrophicness?} (Q by \href{https://english.stackexchange.com/users/23661/solvingpuzzles}{solvingPuzzles})
    
    \item \url{https://english.stackexchange.com/questions/134352/should-i-modify-a-gerund-using-an-adjective-or-an-adverb} (Q by \href{https://english.stackexchange.com/users/55657/worawit-tepsan}{worawit-tepsan})
    
    \item \url{https://english.stackexchange.com/questions/222567/what-is-the-etymology-of-physician} (Q by \href{https://english.stackexchange.com/users/106789/casvaart}{casvaart})
    
    \item \url{https://ell.stackexchange.com/questions/185516/did-vs-have-done} (Q by \href{https://ell.stackexchange.com/users/85088/learner}{learner})

    \item \url{https://english.stackexchange.com/questions/162824/what-is-the-analog-equivalent-of-digitally} (Q by \href{https://english.stackexchange.com/users/59963/rocketmagnet}{rocketmagnet}, first A by \href{https://english.stackexchange.com/users/71639/allisonashley}{AllisonAshley}, second A by \href{https://english.stackexchange.com/users/70861/hot-licks}{Hot Licks})
    
     \item \url{https://ell.stackexchange.com/questions/13749/explain-the-verb-tense-in-all-that-glitters-is-not-gold} (Q by \href{https://ell.stackexchange.com/users/1821/chinmay235}{Chinmay235}, first A by \href{https://ell.stackexchange.com/users/113/j-r}{J.R.}, second A by \href{https://ell.stackexchange.com/users/21609/sajad}{
sajad})

\item \url{https://english.stackexchange.com/questions/162824/what-is-the-analog-equivalent-of-digitally} (Q by \href{https://english.stackexchange.com/users/59963/rocketmagnet}{Rocketmagnet})

\item \url{https://english.stackexchange.com/questions/203518/why-is-this-sentence-additional-nine-features-were-added-incorrect} (Q by \href{https://english.stackexchange.com/users/95069/user95069}{user95069}), A by \href{https://english.stackexchange.com/users/71333/nick2253}{Nick2253}

\item \url{https://english.stackexchange.com/questions/49384/ed-and-ing-adjectives} (Q by \href{https://english.stackexchange.com/users/15191/itun}{itun})

\item \url{https://ell.stackexchange.com/questions/87725/what-does-in-refer-to} (Q by \href{https://ell.stackexchange.com/users/28709/anfi}{Anfi})

\item \url{https://english.stackexchange.com/questions/102996/returning-for-the-nth-time} (Q by \href{https://english.stackexchange.com/users/37080/alicorntwilightisatroll}{AlicornTwilightisaTroll})

\item \url{https://english.stackexchange.com/questions/55331/why-is-this-usage-of-to-incorrect} (Q by \href{https://english.stackexchange.com/users/3383/ademos}{Ademos})

\item \url{https://ell.stackexchange.com/questions/87725/what-does-in-refer-to} (Q by \href{https://ell.stackexchange.com/users/28709/anfi}{Anfi})

\item \url{https://ell.stackexchange.com/questions/322637/he-is-more-than-a-friend-is} (Q by \href{https://ell.stackexchange.com/users/82229/loviii}{Loviii}, first A by \href{https://ell.stackexchange.com/users/146667/marcinmanhattan}{MarcInManhattan}, second A by \href{https://ell.stackexchange.com/users/20756/kirt}{
Kirt})

\item \url{https://english.stackexchange.com/questions/258060/verb-for-doing-something-unknowingly} (Q by \href{https://english.stackexchange.com/users/128480/daniel-bramhall}{Daniel Bramhall
}, first A by \href{https://english.stackexchange.com/users/127726/chasly-supports-monica}{chasly - supports Monica
}, second A by \href{https://english.stackexchange.com/users/78544/talrnu}{
talrnu})

\item \url{https://ell.stackexchange.com/questions/322580/know-someone-in-detail} (Q by \href{https://ell.stackexchange.com/users/150655/simo-ita}{Simo Ita})

\end{itemize}

\section{On our use of z-scores}\label{sec:zscore}

In our human evaluation, raters were presented with a question and five candidate answers and asked to rate each on a scale from 1 to 5 for each of our two criteria (C1 and C2).

Our main goal is to compare the quality of the answers across 5~conditions (3~systems, 2~posts from the site). Raters may have different interpretations of the absolute scales---for example, some raters could be more generous than others overall in terms of the numerical rating, even if they agree on the ranking of systems.

There are several possible ways to factor out this bias. One way is to compute standard scores, a.k.a.~z-scores, for each annotator's distribution of responses on each criterion.
Consider C1: from the ratings of an annotator $a$ we have the empirical distribution 

\[ p^{\text{C1}}_a(y^{\text{C1}}_{i,a} \mid x_i) \]

\noindent where $i$ indexes the items (answers, of which multiple ones may belong to the same question), and likewise for C2. For each of these distributions we fit a normal distribution by computing mean and standard deviation. For an absolute rating $y^{\text{C1}}_{i,a}$, its z-score $z^{\text{C1}}_{i,a}$ is its number of standard deviations above the mean rating for that annotator on that metric (a negative z-score indicates it is below the mean).
Averaging the z-scores for a particular condition, we can see whether annotators tended to rate outputs in that condition with higher or lower scores than the other outputs they saw in the sample.
Note that the z-score computation ignores the grouping of answers from different conditions into questions, so it is not directly measuring annotators' rankings of candidate answers to a particular question.

\section{Further Details}\label{sec:more}

\begin{table}[h]
\centering\small
\begin{tabular}{l c}
\hline
\textbf{Parameter}&\textbf{Value}\\
\hline
Batch Size&8\\
Max. Gradient Updates&75k\\
Max. Input Length&512\\
Max. Output Length&512\\

\hline
\end{tabular}

\caption{T5 hyperparameters used for the Answer Generation task}
\label{tab:hyper}
\end{table}

\begin{table*}\centering\small
\begin{threeparttable}
  \begin{tabular}{l|c|c|c|c|c}
     & ROUGE-1 & ROUGE-2 & ROUGE-L & BLEU Score&BERTScore\\\hline

Set-1 &\textbf{0.303}&\textbf{0.084}& \textbf{0.201} &\textbf{0.092} & \textbf{0.859}\\
Set-2 &0.296 &0.079 &0.192 &\textbf{0.092}&0.854 \\
Set-3 &0.286 &0.071 &0.193 &0.052&0.856 \\
Set-4 &0.273 &0.067 &0.188 & 0.039&0.857\\
Set-5 & 0.289& 0.074&0.193 &0.051&0.856 \\\hline

  \end{tabular}
  \end{threeparttable}

  \caption{GPT-3 few-shot scores on a subset of dev when using different sets of in-context examples}
  \label{tab:gpt3-dev}
\end{table*}

\begin{table*}\centering
\begin{threeparttable}
  \fontsize{7.3}{9.3}\selectfont
  \begin{tabular}{@{}p{0.15\textwidth} | p{0.8\textwidth}}
  \hline
     \textbf{Title}&\textbf{Question} \\\hdashline
     "He is more than a friend is" & \textbf{(1a)} \textit{He is taller than his friend}. — correct \textbf{(1b)} \textit{He is taller than his friend \textbf{is}}. — as I know, (1b) is also correct, although less common than (1a) \textbf{(2a)} \textit{He is more than a friend}. — correct  \textbf{(2b)} \textit{He is more than a friend \textbf{is}}. — by analogy with (1b), (2b) must be correct too. Is it so? If not, then why is (2b) not correct, whereas (1b) that is analogous to (2b) is correct?\\\hline

     \textbf{Source}& \textbf{Answer} \\\hline

    Top-rated human& This is a bit tricky, because "than" can be a conjunction or a preposition, and not everyone agrees on when it is each. I'll present one interpretation that I believe is pretty standard. In 1a and 1b, "than" is best understood as a conjunction. It compares two actions ("he is taller" and "his friend is tall"). That is why we can insert "is" at the end of 1b. In 2a and 2b, "than" is best understood as a preposition. The prepositional phrase "than a friend" modifies "more". It does \textit{not} compare two actions ("he is more" and "a friend is [adjective]"). That is why we can not insert "is" at the end of 2b. To summarize: In your sentences 1a and 1b, "than" is a conjunction. In 2a and 2b, it is a preposition. Thus, the situations are not analogous. \\\hdashline
    
     Low-rated human& (1a) He is taller than his friend. — correct, but incomplete, and natural (1b) He is taller than his friend is. — correct, but incomplete, and natural (1c) He is taller than his friend is tall. - correct and complete, but awkward (2a) He is more than a friend. — correct, but incomplete, and natural (2b) He is more than a friend is. - correct, but incomplete, and awkward (2c) He is more than a friend to me. - correct, but incomplete, and natural (2d) He is more than a friend is to me. - correct, but incomplete, and awkward (2e) He is more to me than a friend is to me - correct, complete, and awkward The complete statement of meaning is grammatically correct but may be more, or longer, than people naturally say. Typically words that can be implied are eliminated for more natural speech. Some eliminations of implied words will lead to constructions that are grammatically correct and natural, while other eliminations of implied words can lead to constructions that are grammatically correct but which sound awkward. The awkwardness of an expression has more to do with common usage than it does with rules of grammar.\\\hdashline
     
     GPT-3 FS& No, (2b) is not correct. The reason is that (2a) is a comparison between two things, while (2b) is not. In (2a), the comparison is between \textit{he} and \textit{a friend}, while in (2b) the comparison is between \textit{he<} and \textit{his friend is}. This is not a valid comparison, so (2b) is not correct.\\\hdashline
     
     GPT-3 FT-1000& The difference between (1a) and (1b) is that (1a) is a statement about \textit{the friend}, whereas (1b) is a statement about \textit{the friendship}. Consider the following two sentences: <blockquote> John is taller than his friend Bob. </blockquote> vs. <blockquote> John is taller than Bob is. </blockquote> The first sentence is a statement about John and Bob. The second sentence is a statement about John and the friendship between John and Bob. In the first sentence, the friend is the subject of the sentence. In the second sentence, the friend is the object of the sentence. In your examples, (1a) and (2a) are statements about the friend, whereas (1b) and (2b) are statements about the friendship.\\\hdashline
     
     T5-xxl &  Answering your question, I can say that, no, (2b) is not correct. (2a) "He is taller than his friend" is idiomatic and natural, and the simple present tense is used here to describe an action or state. (2b) "He is more than a friend" is not idiomatic and does not mean the same thing.\\

    \hline

    \textbf{Title}&\textbf{Question} \\\hdashline
     Verb for doing something unknowingly& I cannot think of an effective verb that would suggest someone is doing something unknowingly yet doing it nonetheless - almost like acquiescing. I have thought of 'sleepwalking' however there must be something better. For example, VERB into a materialistic society.\\\hline

     \textbf{Source}& \textbf{Answer} \\\hline
    Top-rated human&They are \textbf{drifting} into a materialistic society. They are \textbf{sliding} into a materialistic society.\\\hdashline
    
     Low-rated human& \textit{Evolve} fits your needs and in particular your example pretty closely. Evolution is a natural process, implying the lack of intention you're looking for. It also fits your example as it's an act of change and development. Similarly, you might consider \textit{mutate}, another natural occurrence with greater emphasis on randomness and less emphasis on a direction of development. More generally, you could say my answer is: use a context-specific verb which emphasizes nature (i.e. the lack of human intention) as the driving force behind the action.\\\hdashline
     
     GPT-3 FS& The verb \textit{drift} could be used in this context. It suggests a lack of conscious direction or control, and implies that the person is being carried along by external forces. <blockquote> He \textit{drifted} into a materialistic society.</blockquote> \\\hdashline
     
     GPT-3 FT-1000&I think \textit{bemused} would work well in this context. It means to confuse or bewilder, but can also be used to describe someone who is behaving in a confused way, or who is reacting to something confusing. So, "Jane was bemused by the materialistic society she found herself in." or "Jane was bemused by her own behaviour as she bought yet another pair of designer shoes.\\\hdashline
     
     T5-xxl&  How about  \textit{slumbering} into a materialistic society?
 \\

    \hline
    
    \end{tabular}
  
  \end{threeparttable}

  \caption{Two example posts with human and model-generated answers.}
  \label{tab:sample-gen-a-2}
\end{table*}

\begin{figure*}[]
\centering
\minipage{0.5\textwidth}
\centering
  \includegraphics[scale=.4,trim = 3cm 8cm 3cm 8cm, clip]{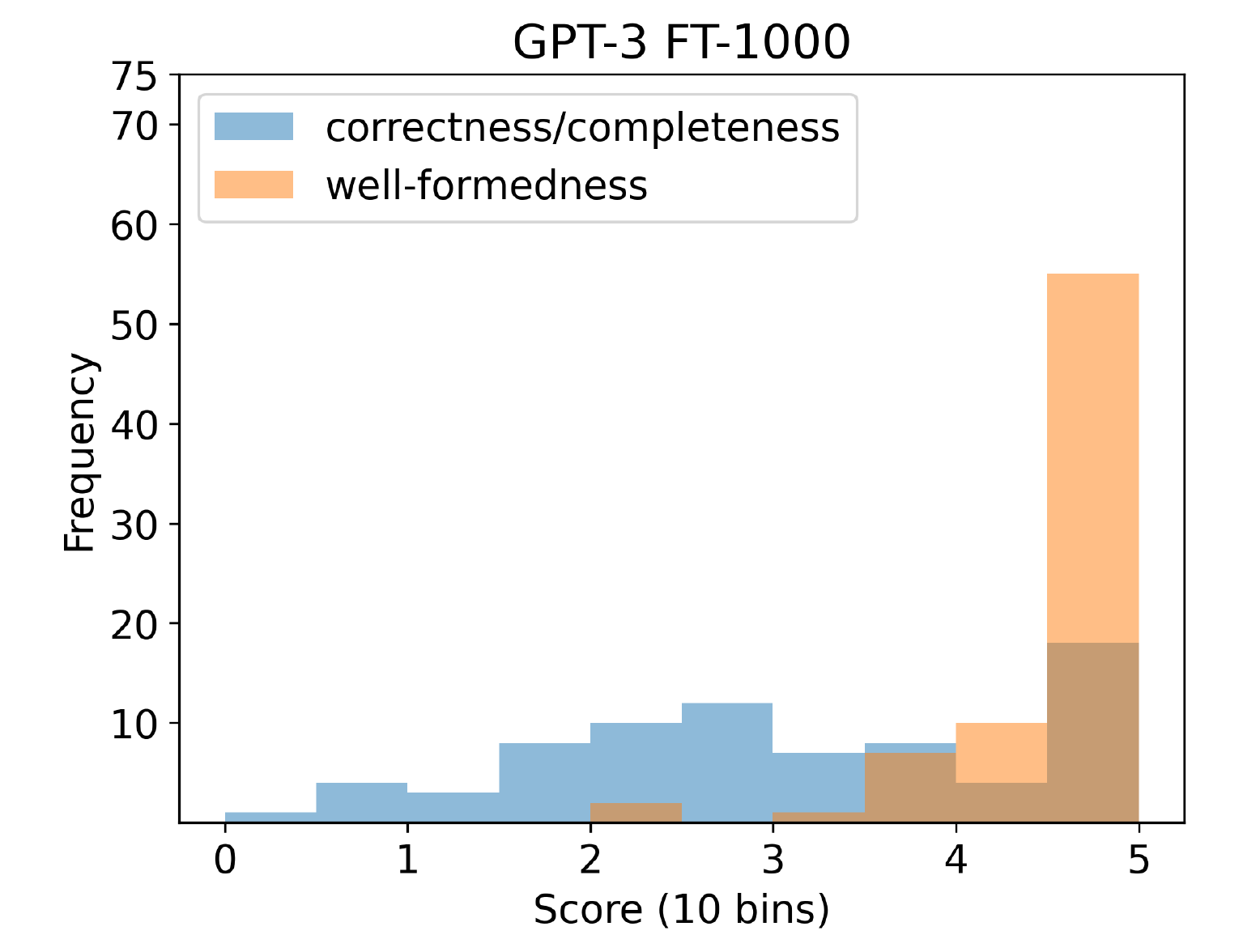}
  
\endminipage\hfill
\minipage{0.5\textwidth}
\centering
  \includegraphics[scale=.4,trim = 3cm 8cm 3cm 8cm, clip]{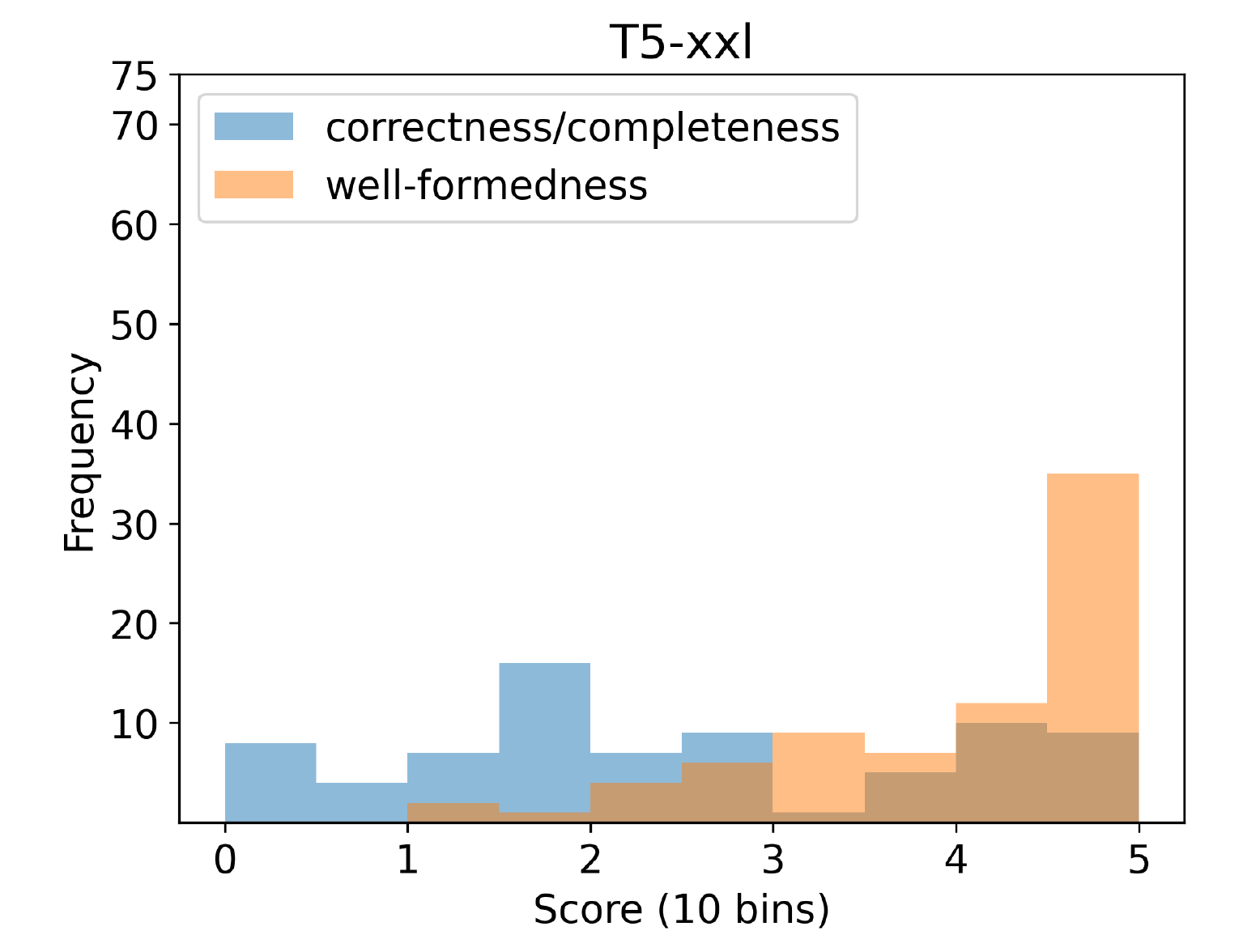}
\endminipage\hfill

\caption{Histograms of the average score of each two annotators from our human evaluation for answers generated by GPT-3 FT-1000 and T5-xxl. The other answers evaluated are represented in \cref{fig:human_charts}.}\label{fig:more_human_charts}

\end{figure*}

\begin{figure*}
    \centering
        \includegraphics[width=0.8\textwidth ,trim = 0cm 0cm 0cm 0cm, clip]{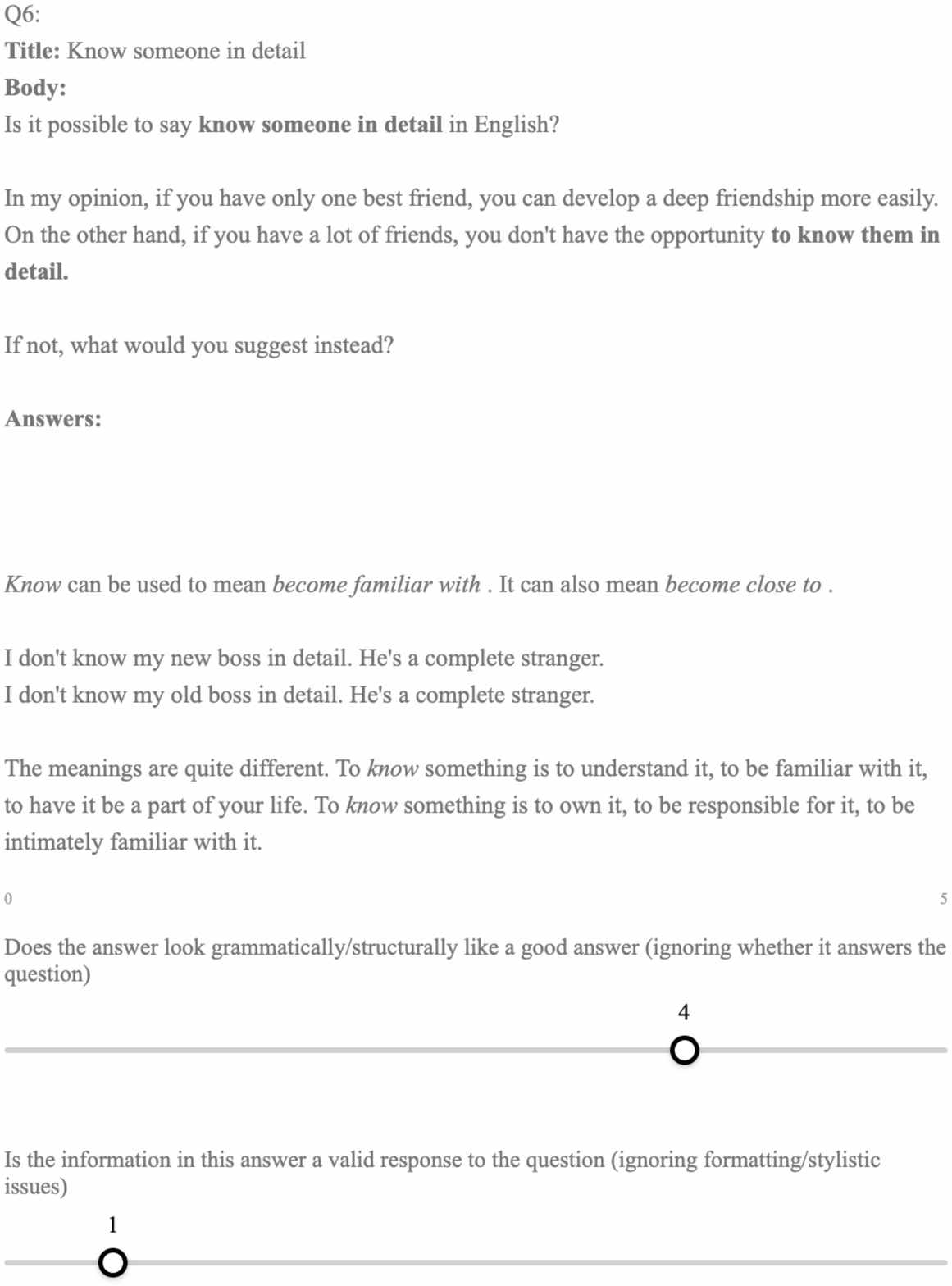}
        \caption{Screenshot from the survey we designed for human evaluation on the Qualtrics platform. Five answers were shown for each question as described in \cref{sec:gen}.}
        \label{fig:survey}
\end{figure*}

\end{document}